\documentclass[10pt, a4paper]{article}
\usepackage{lrec2022}
\usepackage{graphicx}
\usepackage{tabularx}
\usepackage{microtype}
\usepackage{booktabs}
\usepackage{enumitem}
\newlist{compactdesc}{description}{1}
\setlist[compactdesc, 0]{labelindent=0ex, labelsep=0pt, nosep, style=unboxed,leftmargin=0mm}
\usepackage[utf8]{inputenc}
\usepackage{tipa}
\usepackage[hidelinks]{hyperref}
\usepackage[dvipsnames]{xcolor}
\usepackage{xspace}
\usepackage{subfig}
\usepackage{wrapfig}

\newcommand{\emotionname}[1]{\textit{#1}}
\newcommand{\fear}{\emotionname{fear}\xspace}
\newcommand{\joy}{\emotionname{joy}\xspace}
\newcommand{\anger}{\emotionname{anger}\xspace}

\newcommand{\guilt}{\emotionname{guilt}\xspace}
\newcommand{\shame}{\emotionname{shame}\xspace}
\newcommand{\hope}{\emotionname{hope}\xspace}

\newcommand{\sadness}{\emotionname{sadness}\xspace}

\newcommand{\disgust}{\emotionname{disgust}\xspace}
\newcommand{\pleasantness}{\emotionname{pleasantness}\xspace}

\newcommand{\effort}{\emotionname{effort}\xspace}
\newcommand{\understand}{\emotionname{understand}\xspace}

\newcommand{\adjCheck}{\emotionname{adjustment check}\xspace}
\newcommand{\discrepancy}{\emotionname{expectation discrepancy}\xspace}
\newcommand{\otherResp}{\emotionname{other responsibility}\xspace}
\newcommand{\certainty}{\emotionname{certainty}\xspace}
\newcommand{\novelty}{\emotionname{novelty}\xspace}
\newcommand{\goal}{\emotionname{goal relevance}\xspace}
\newcommand{\attention}{\emotionname{attention}\xspace}
\newcommand{\selfResp}{\emotionname{self responsibility}\xspace}
\newcommand{\responsibility}{\emotionname{responsibility}\xspace}
\newcommand{\suddenness}{\emotionname{suddenness}\xspace}
\newcommand{\selfControl}{\emotionname{self control}\xspace}
\newcommand{\situationalControl}{\emotionname{situational control}\xspace}

\newcommand{\urgent}{\emotionname{urgency}\xspace}

\newcommand{\F}{$\textrm{F}_1$\xspace}

\makeatletter
\def\blfootnote{\xdef\@thefnmark{}\@footnotetext}
\makeatother

\newcommand{\corpusname}{x-en\textsc{Vent}\xspace}
\title{\corpusname: A Corpus of Event Descriptions with Experiencer-specific\\
  Emotion and Appraisal Annotations}

\name{Enrica Troiano$^*$, Laura Oberl\"ander$^*$, Maximilian Wegge, Roman Klinger} 

\address{Institut f\"ur Maschinelle Sprachverarbeitung \\
  University of Stuttgart \\
  \{enrica.troiano,laura.oberlaender,maximilian.wegge,roman.klinger\}@ims.uni-stuttgart.de\\
}

\abstract{%
  Emotion classification is often formulated as the task to categorize
  texts into a predefined set of emotion classes. So far, this task
  has been the recognition of the emotion of writers and readers, as
  well as that of entities mentioned in the text. We argue that a
  classification setup for emotion analysis should be performed in an
  integrated manner, including the different semantic roles that
  participate in an emotion episode. Based on appraisal theories in
  psychology, which treat emotions as reactions to events, we compile
  an English corpus of written event descriptions. The descriptions
  depict emotion-eliciting circumstances, and they contain mentions of
  people who responded emotionally.  We annotate all experiencers,
  including the original author, with the emotions they likely
  felt. In addition, we link them to the event they found salient
  (which can be different for different experiencers in a text) by
  annotating event properties, or \textit{appraisals} (e.g., the
  perceived event undesirability, the uncertainty of its outcome).
  Our analysis reveals patterns in the co-occurrence of people's
  emotions in interaction. Hence, this richly-annotated resource
  provides useful data to study emotions and event evaluations from
  the perspective of different roles, and it enables the development
  of experiencer-specific emotion and appraisal classification
  systems.  \\[\baselineskip]%
}

\begin{document}

\maketitleabstract

\blfootnote{\hspace{-0.65cm}$^*$The first two authors contributed
  equally.}

\section{Introduction}

Computational emotion analysis from text includes various subtasks,
with the (arguably) most popular one being emotion classification --
i.e., to categorize texts into a predefined set of emotion classes
\cite{Mohammad2018}.  The adopted categories often coincide with the
list of ``basic emotions'' proposed by \newcite{Ekman1992}, namely
fear, joy, sadness, anger, disgust, surprise, or with inventories
defined by other theorists, like \newcite{Plutchik2001}.  In addition
to discrete labels, some works have been identifying structured
information, namely emotion roles, which divides texts into spans that
correspond to semantic roles. Depending on the considered domain, it
can aim at distinguishing an emotion experiencer, a stimulus (i.e.,
the triggering event), a target towards which the emotion is directed,
or a cue word evoking a specific affective state \cite{Mohammad2014}.

Structured knowledge of this sort has been proven informative for
emotion classification \cite{Oberlaender2020b}, but to face one
challenge at a time, the two strands of research are typically
addressed separately.  Hence, when the emotion experiencer is not
known to a classification model, a preliminary decision has to be
made: the classification should regard either the emotion expressed by
the writer \cite[i.a.]{mohammad-2012-emotional}, or the one triggered
in the reader of the text \cite[i.a.]{Haider2020}. Studies that
include both perspectives are rare
\cite[i.a.]{Bostan2020,buechel-hahn-2017-readers}, and so are those
focused on the emotions of characters mentioned in the text
\cite{Kim2019,Kim2018}. In fact, they are mainly dedicated to
\begin{wrapfigure}[8]{r}{0.3\textwidth}
\vspace{5pt}
  \centering
  \includegraphics[scale=0.7]{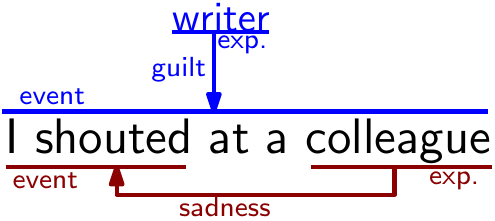}
  \caption{Stimulus and emotion annotation specific to experiencers.}
  \label{fig:example}
\end{wrapfigure}
the
analysis of emotion in literature -- a task that comes
with its own challenges, due to the artistic nature of the domain.
Our paper sets a different focus: to account for a more complete
understanding of an utterance's affective content, we are interested
in the perspective of diverse emotion experiencers, both the writer of
an event description and other in-text entities.

We compile \corpusname,\footnote{We call the corpus \corpusname, to
  indicate that it is English and annotated by trained
  e\textsc{x}perts, in contrast to a crowdsourced resource under
  development (Crowd-en\textsc{Vent}).}, a corpus
predominantly made of self-reported event descriptions. The texts were
written by people who felt a particular emotion in response to such
events, many of which involved third parties
\cite{Troiano2019}. Hence, (1) we mark the textual spans corresponding
to the emotion-triggering events and all of their experiencers (the
writer being a special token), (2) we draw relations between the two,
and (3) we specify what emotion results from their interplay.  An
example is shown in Figure~\ref{fig:example}, where the writer feels
guilty for their own behaviour, while the person affected by it more
likely feels sadness.

These $\langle$event, experiencer, emotion$\rangle$ tuples reflect the
model of \textit{appraisal} \cite{scherer1989appraisal}, according to
which emotions arise in response to the cognitive evaluation of
events.  This psychological theory places emotions in a
multidimensional space: each dimension represents a specific property
of the event being evaluated, for instance its perceived \pleasantness
(likely to be high with joy and low with disgust), the mental or
physical \effort that it can be expected to cause in the experiencer
(likely high with episodes of anger or fear), the \responsibility held
by the experiencer for what has happened (high for guilt)
\cite{Smith1985}.  The annotation unit that we present in this work
therefore consists of
$\langle \textrm{ev}, \textrm{exp}, \textrm{emo},
\overrightarrow{\textrm{appraisal}}\rangle$ tuples, in which the event
(ev) and the experiencer (exp) are token spans, the emotion (emo) is a
single emotion label, and $\overrightarrow{\textrm{appraisal}}$ is a
vector of numeric values for various appraisal dimensions.

Hence, we bring together work on appraisal theories for text analysis
\cite{balahur2011building,Hofmann2020,Hofmann2021}, emotion role
labeling \cite{Oberlaender2020b,Mohammad2014,Kim2018}, and emotion
classification. We combine the annotation layers, as exemplified in
Figure~\ref{fig:example}, an example where the dimension of
\responsibility is scored with the highest degree for the writer and
lowest for ``a colleague'', while \pleasantness is low for both.  Our
corpus encompasses 720 event descriptions\footnote{Available at
  \url{http://www.ims.uni-stuttgart.de/data/appraisalemotion}} and
enables the development of experiencer-specific emotion and appraisal
analysis systems. It further enables analyses of the interplay of
emotions of people in interaction, as it emerges from text.

\section{Previous Work}
\paragraph{Resources for Emotion Recognition.}
The construction of emotion resources typically relies on
psychological models of emotions. Following theories of basic emotions
\cite{Ekman1992,Plutchik2001}, texts can be labelled with categorical
classes.  Alternatively, emotions can be described via continuous
values in a vector space.  Such is the basis of studies like
\newcite{Preotiuc2016},\newcite{Yu2016} and \newcite{Buechel2017},
which comprise the dimensions of valence, arousal and dominance
motivated by \newcite{russell1977evidence}.

Usually, emotion classification and resource construction associate a
text to one or more emotions from a specific perspective. Indeed,
emotions arise in language whenever writers mention or evoke a mental
state of their own or that of others (e.g., a character), as well as
when they attempt to elicit a reaction in their readers. This has
motivated the design of corpora with texts from various domains, like
Reddit comments \cite{demszky-etal-2020-goemotions}, tales
\cite{Ovesdotter2005}, blogposts \cite{aman2007identifying}, and
labelled with the emotion of writers \cite{mohammad-2012-emotional},
of readers \cite{chang-etal-2015-linguistic}, or of both
\cite{buechel-hahn-2017-readers}.

As opposed to these studies, we annotate event descriptions from the
perspective of each experiencer mentioned or presupposed (i.e., the
writer) in the text.

\paragraph{Structured Emotion and Sentiment Analysis.} 

Our setup is close to previous work in structured sentiment
analysis. There, opinion holders are extracted
\cite{Toprak2010,Kessler2010}, along with ``aspects'' and sentiment
polarity values revealing the relation between aspect and
holder. However, the linguistic variability of descriptions of
emotion-inducing events is comparably richer than sentiment opinion
expressions \cite{Klinger2014}: not only the experiencer needs to be
situated in a given circumstance, but the link between such
circumstance and the consequent emotion is to be grasped via world
knowledge (e.g., that shouting at somebody, like in
Figure~\ref{fig:example}, might be inappropriate).

Structured emotion analysis, on its part, has aimed at identifying
segments of texts that mention emotion experiencers or stimuli
\cite{wei-etal-2020-effective,Neviarouskaya2013}.  Accordingly, the
available resources contain labels at the sub-sentence
level. \newcite{gao2017overview}, for instance, built a corpus which
marks emotion cause segments; \newcite{ghazi2015detecting} did the
same by leveraging emotion frames in FrameNet \cite{fillmore1976frame}
that include a \textit{stimulus} argument. \newcite{Oberlander2020}
compared clause-level and token-level stimulus detection.

Similar to corpora on emotion stimulus detection
\cite{russo-etal-2011-emocause,gui-etal-2016-event,li2014text,xia-ding-2019-emotion,chen-etal-2020-end,Kim2018,Bostan2020,Mohammad2014},
we consider emotion causes, or stimulus events, but we extend our
definition of experiencers to both writers and third entities.  We
point out the ways in which they appraise events and the resulting
emotion reaction, which is reconstructed from (but not felt by) the
annotators.

\paragraph{Events and Appraisals.}

Appraisal annotations have enlivened some research efforts so far.
\newcite{Hofmann2020} exploited the idea that appraisals enable
readers to interpret what others feel. They set up an annotation task
in which event descriptions coming from en\textsc{Isear}
\cite{Troiano2019}, already labelled with the emotions of their
writers, were associated to seven appraisal dimensions.  Using the
same dimensions and corpus, \newcite{Hofmann2021} experimented with
different annotation strategies.  Compared to our work, they have
disregarded the multitude of emotion perspectives available in text
and only considered a limited number of appraisal dimensions.

Other than corpora, the knowledge base EmotiNet was motivated by
appraisal theories \cite{balahur2011building}.  It describes events
with respect to their atomic elements, such as actors, actions and
objects, as well as their properties, defined along the lines of
appraisal criteria.  Further, \newcite{Cambria2020} presented a
logical representation of events inspired by appraisal theories, but
performed sentiment analysis, and \newcite{Shaikh2009} used logical
expressions to combine event properties with the goal to infer an
emotion category.

\newcommand{\statement}[1]{\multicolumn{2}{l}{#1}\hfill{}1--5\hspace{5mm}}
\begin{table*}
  \centering\small
  \renewcommand{\arraystretch}{0.89}
\begin{tabular}{llp{35mm}}
  \toprule
  Variable & Description & Values\\
  \cmidrule(r){1-1} \cmidrule(r){2-2}\cmidrule(r){3-3}
  Event & the most salient fact for the evaluation of an emotion experience & span\\
  \cmidrule(r){1-1} \cmidrule(r){2-2}\cmidrule(r){3-3}
  Experiencer(s) & the person(s) involved in the situation, and aware of it & span\\
  \cmidrule(r){1-1} \cmidrule(r){2-2}\cmidrule(r){3-3}
  Emotions & discrete names representing responses to events & disgust, joy, guilt, hope, sadness, surprise, shame, trust \\
  \cmidrule(r){1-1} \cmidrule(r){2-2}\cmidrule(r){3-3}
  Appr. Dimensions: & &\\
 \textit{- suddenness} & \statement{the event was sudden or abrupt}\\
  \textit{- familiarity} & \statement{the event was familiar to the experiencer}\\
  \textit{- pleasantness} & \statement{the event was pleasant for the experiencer}\\
  \textit{- understand} & \statement{the experiencer understood what was happening}\\
  \textit{- goal relevance} & \statement{the event was important or relevant for experiencer's goals}\\
  \textit{- self responsibility} & \statement{the event was caused by experiencer's own behaviour}\\
  \textit{- other responsibility} & \statement{the event was caused by somebody else's behaviour}\\
  \textit{- situational respons.} & \statement{the event was caused by chance or special circumstances}\\
  \textit{- effort} & \statement{the situation required the experiencer a great deal of energy}\\
  \textit{- exert} & \statement{the experiencer felt they needed to exert themselves to handle the event}\\
  \textit{- attend} & \statement{the experiencer had to pay attention to the situation}\\
  \textit{- consider} & \statement{the experiencer wanted to consider the situation}\\
  \textit{- outcome probability} & \statement{the experiencer could anticipate the consequences of the event}\\
  \textit{- expect.\ discrepancy} & \statement{the experiencer did not expect that the event would occur}\\
  \textit{- goal conduciveness} & \statement{the event itself was positive or it had positive consequences for the experiencer}\\
  \textit{- urgency} & \statement{the event required an immediate response from the experiencer}\\
  \textit{- self control} & \statement{the experiencer had the capacity to affect the event}\\
 \textit{- other control} & \statement{someone or something other than the experiencer was influencing what was going on}\\
  \textit{- situational control} & \statement{the situation was the result of outside influences of which nobody had control}\\
  \textit{- adjustment check} & \statement{the experiencer anticipated that they could live with the consequences of the event}\\
  \textit{- internal check} &\statement{the event clashed with the experiencer's ideals and standards}\\
  \textit{- external check} & \statement{the event violated laws or social norms}\\
  \bottomrule
\end{tabular}
\caption{The variables in our annotation task: event, experiencers, emotions, and 22 appraisal dimensions.}
\label{tab:annotation-variables}
\end{table*}

\section{Appraisals}
Appraisal theories approach emotions as componential processes, that
is, as ``an episode of interrelated, synchronized changes in the
states of all or most of the five organismic subsystems in response to
the \textit{evaluation} of a [...]  stimulus-event''
\cite{Scherer2005}. The five subsystems are the cognitive (the appraisal),
neurophysiological (bodily symptoms), motivational (action tendencies)
and motor components (facial and vocal expressions), as well as a component related to
the subjective feelings (the perceived emotional experience).  All of them
have corresponding linguistic realizations that evoke an emotion --
e.g., bodily symptoms $\rightarrow$ ``he couldn't stop shaking''
$\rightarrow$ fear \cite{Casel2021}, but the cognitive component plays
an additional role for emotion analysis, as it enables emotion
decoding.  Humans' empathy and ability to take the affective
perspective of others is guided by the assessment of whether a certain
event might have been important, threatening, or convenient for those
who lived through it \cite{omdahl1995cognitive}.

The change in appraisal hence consists in evaluating the situation
with respect to the significance it holds for an individual: Does the
current event hamper my goals? Can I predict what will happen next?
Do I care about it?  While the criteria that are used to assess a
situation can in principle be countless, appraisal theorists have come
up with a number of criteria contributing to the development of an
emotional episode \cite[among
others]{Ellsworth1988,Roseman1990,Tracy2006}.  Most of them comprise
the appraisal of \goal.  Others include \pleasantness and \novelty.
In our study, we use 22 appraisal dimensions (see
Table~\ref{tab:annotation-variables}), based on \newcite{Smith1985},
\newcite{Scherer1997} and \newcite{Scherer2013}.

These studies have developed a set of questions (e.g., ``did you think
that the event was pleasant?'')  in order to collect self-reports on
appraisals.  \newcite{yanchus2006development} raised concerns that
this wording might bias the respondents: questions give people the
chance to develop a theory in retrospect about their behaviour;
instead, statements leave participants free to recall if the depicted
behaviours (e.g., ``I thought the event was pleasant''), applied to
them or not.  We abide by this idea and spell out each appraisal as an
affirmation.

Appraisals hence reveal one's interaction with the environment.  Two people
with different goals, cultures and sets of beliefs might produce
different evaluations of a stimulus. Consequently, specific appraisal
combinations lead to different emotion reactions. \newcite{Smith1985},
for instance, qualify 15 emotions on the basis of \pleasantness,
\responsibility of the emotion experiencer for triggering the event,
\certainty about what was happening, \attention put on the emotion
stimulus, \effort expended to deal with it, and \situationalControl,
or the ability to influence the development of the situation.  This
can account for differences in how people respond to an event, as well
as differences in the emotion inferred from (and chosen for) a text by
annotators.

\begin{figure*}
  \centering
  \includegraphics[scale=1.2]{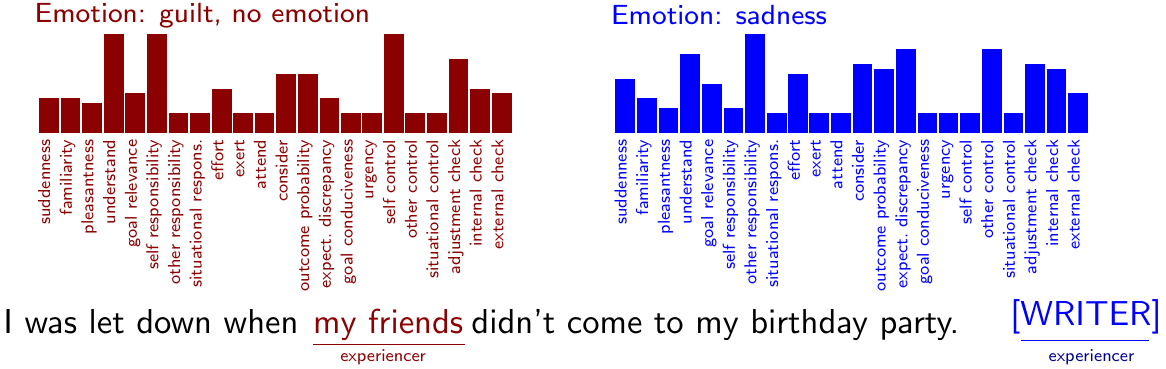}
  \caption{Annotated example. The event that is linked to both
    experiencers corresponds to the whole text here.}
  \label{fig:annotation-example}
\end{figure*}

\section{Corpus Creation}
Our goal is to populate a corpus with appraisal dimension ratings
for each experiencer mentioned in the text, quantifying the degree to
which each property applies to a given situation. Tracing back the
original text authors to obtain first-hand appraisal
self-recollections would be unfeasible. For this reason, we relied on
the help of external readers.  Their task consisted in assigning
emotion labels, appraisal dimension ratings, and span annotations
indicating experiencers and events in each text.  Specifically, we
asked our annotators to put themselves in the shoes of each
experiencer to reconstruct both the emotion they felt and the way in
which they might have appraised the event.

We conducted the annotation on the platform
I\textsc{nce}p\textsc{tion}\footnote{\url{https://inception-project.github.io}}
\cite{tubiblio106270}. Annotators were four master students of
computer science and computational linguistics, three male and one
female, aged between 24 and 28. They were familiar with the field of
emotion analysis and with appraisal theories, but the guidelines for
the task still provided them with extensive examples and definitions
for each concept to be annotated.

\subsection{Data}
The instances in our resource were sampled from various corpora that
contain event descriptions and emotion annotations -- the latter
mostly provided by the authors of the texts themselves.

During the training phase of the annotators, we extracted data from
the \textsc{Isear} corpus produced by psychologists
\cite{scherer1994evidence,Scherer1997}, from its kin en\textsc{Isear}
\cite{Troiano2019}, \textsc{Empathetic-Dialogues}
\cite{rashkin-etal-2019-towards} and Event2Mind
\cite{rashkin-etal-2018-event2mind}.  Later, we extracted texts only
from en\textsc{Isear}, which spans 1001 English sentences describing
real-life events associated to 7 emotions.\footnote{Our corpus
  predominantly consists of data from enISEAR.} Emotion names were
manually masked by the authors of the corpus, such that follow-up
emotion interpretation tasks based on those instances would not result
in a trivial endeavour.  Therefore, the texts coming from
en\textsc{Isear} are implicit emotion expressions, in which the
affective meaning of texts is evoked (e.g., ``I felt ... when my
grandad passed away'', ``I felt ... when I first flew on a plane''),
rather than spelled out.

\subsection{Annotation Guidelines}
Annotators were presented with one instance at a time.  The first step
they had to accomplish was to assess whether the text contained an
event. If they spotted one, they performed the following:

\begin{compactdesc}
\item \textbf{1. experiencer span identification}, aimed at marking the
  textual span that contains the experiencer mention;
\item \textbf{2. salient events span identification}, i.e., marking the
  portion of text containing the event appraised by such
  experiencer\footnote{Experiencer and event spans can overlap.};
\item \textbf{3. emotion selection}, to choose the reaction that the
  experiencer most likely had to the appraised event;
\item \textbf{4. appraisal dimension rating}, which consisted in scoring
  the value of each appraisal dimension with respect to the event.
\end{compactdesc}

Annotators repeated these steps for each event and experiencer.  An
example of the resulting annotation is provided in
Figure~\ref{fig:annotation-example}.

\subsubsection{Experiencer Span Identification}
A text might mention different entities, but not all of them should
automatically be deemed experiencers. The guidelines characterized an
experiencer as the person who is involved in the situation, is aware
of it, evaluates it, and is somewhat affected by what happened. In
``Helen didn't notice that Julia lost her keys'', only ``Julia'' would
be the experiencer, while both would be annotated in ``Julia lost
Helen's keys, but Helen wasn't bothered and kept focusing on her
homework'', in which they likely react in different ways (e.g., Julia
$\rightarrow$ guilt, Helen$\rightarrow$no emotion).  This example
shows that (1) experiencers do not necessarily feel an emotion, (2)
experiencers can include multiple entities, (3) each of them can be
linked to different events (e.g., losing the keys, focusing on
homework). Considering if experiencers are aware of the event is key
to determine if they appraised the event but felt no emotion or were
present in an event but did not assess it -- irrelevant in our setup.

Experiencers could be separately marked in the text (i.e., ``Julia'',
``Helen'' individually), or be considered as a unique entity if
the two were associated to the same emotion, elicited by the same
event (e.g., ``my friend and Helen passed the exam'').  Annotators
were instructed to select the {[}WRITER{]} token, appended at
the end of each instance, if they judged the text's author to be an
experiencer.  In ``my daughter was building a snowman'', ``my
daughter'' would be annotated along with {[}WRITER{]}, whose
involvement can be recognized by the possessive ``my'', while in
``Mary was building a snowman'' only ``Mary'' would be taken as an
experiencer.

\subsubsection{Salient Events Span Identification}
We gave a loose definition of ``event'', qualifying it as the occasion
or the happening that is the most relevant for the evaluation
(appraisal) of the experience. For instance, despite sharing much
lexical material, the sentences ``I attended the funeral of my
grandma'' and ``They started to yell at the funeral of my grandma''
can be said to contain different events (i.e., the funeral vs.\ the
yelling) that took the focus of their experiencer and that, for this
very reason, can be considered salient.  In other words, a salient
event is one which is evaluated by its participants.

We invited the annotators to include the arguments of predicates when
marking an event (e.g., with transitive verbs, the event should
include the object).  Moreover, if a sentence contained different
events and one experiencer (e.g., ``I just bought a brand new car. I
let my brother drive it even though he isn't a good driver''), only
the focal event of the evaluation, which eventually elicited the main
emotion, was annotated.

\subsubsection{Emotion Selection}
For an experiencer and an event, one label could be chosen among:
\textit{anger}, \textit{disgust}, \textit{fear}, \textit{guilt},
\textit{hope}, \textit{joy}, \textit{sadness}, \textit{shame},
\textit{surprise}, \textit{trust}, \textit{disappointment},
\textit{frustration}, \textit{anticipation}, \textit{contentment}, or
\textit{pride}.  Annotators could indicate that the inferred emotion
did not fit any using the option ``other'', or signal the absence of
an emotion reaction by picking the label ``no
emotion''.\footnote{Events can be appraised without leading to an
  emotion.} Note that the data under consideration came with a prior
emotion distribution and prevalent categories. We chose a richer set
of emotions than in the enISEAR data, because we did not want to limit
the possible emotions for other mentions of experiencers than the
writer.

\subsubsection{Appraisal Dimension Rating}
Annotators aimed at reconstructing how events were assessed by
experiencers relative to the 22 dimensions in
Table~\ref{tab:annotation-variables}.  The rating was done on a 1-to-5
scale: the score given to a dimension represents the degree to
which (according to the annotators) the event experiencer would
agree with the statement describing the appraisal.

\subsection{Data Aggregation}
\label{sec:data-aggregation}
As each instance was labeled by four annotators, we aggregate their
decisions into one final adjudicated annotation.  The gold span-level
annotations (experiencers and events) consist in the overlap between
the majority of annotators' decisions, that is, in the shortest span
appearing in all annotations. For instance, with individual
annotations being ``my friend'', ``my friend'', ``my friend and I'',
and ``friend'', the aggregated annotation would be ``friend''. In case
the annotators' decisions differ considerably, or there is no
token-level majority vote, we make use of a combination of automatic
and heuristics-based manual checks, and aim at a high-recall approach
including all entities involved in the emotion episode.  Using the
above example: if two annotators marked the whole phrase ``my friend
and I'' as a single experiencer, another did not find any event
experiencer, and the last only chose the token {[}WRITER{]}, we would
manually align the personal pronoun ``I'' to {[}WRITER{]}, and
propagate to it the emotion and appraisal annotations.  Hence, we
would end up with two experiencers: ``my friend'' and
{[}WRITER{]}. 184 instances required this manual intervention.

Once the experiencers are defined, we aggregate the emotion
annotations and appraisals. For the former, we include the disjunction
of all emotion labels by all annotators who labeled such overlapping
entities (e.g., all emotions associated to ``my friend'', ``my friend
and I'', and ``friend''). For the latter, we aggregate the ratings to
a minimum, maximum, and average score. Note that in the published
corpus, we provide the original individual annotations in addition.

\begin{table}
  \centering\small
  \begin{tabular}{lccc}
    \toprule
    Variable    & Exact-F$_1$ & Relaxed-F$_1$ & Cohen's $\kappa$ \\
    \cmidrule(r){1-1}\cmidrule(rl){2-2}\cmidrule(l){3-3}\cmidrule(l){4-4}
    Experiencer & 0.86 & 0.88 & 0.84 \\
    Event       & 0.34 & 0.86 & 0.75\\
    Emotion    & -- & -- & 0.62\\
    \bottomrule
  \end{tabular}
  \caption{Inter-Annotator agreement for span
    annotations and emotion category. Scores for individual
    emotions together with frequencies are
    in Table~\ref{tab:iaa-emo}.}
  \label{tab:iaa-emo-exp-event}
\end{table}

\section{\corpusname Analysis}

\subsection{Inter-Annotator Agreement}
Inter-annotator agreement results are in
Table~\ref{tab:iaa-emo-exp-event}.  We show three measures. In
Exact-F$_1$, we take one annotator as the gold standard and the other
as a prediction -- repeating this procedure for each annotator pairs.
An annotation span is accepted as a true positive if the whole span is
exactly matching the other annotator's span. In the variant
Relaxed-F$_1$, we accept a true positive if there is at least a
one-token overlap between the two annotator's spans. We report
averages across all annotator pairs.  Lastly, Cohen's $\kappa$
\cite{cohen1960coefficient} refers to the average of a token-level
pairwise assessments, in line with standard corpus annotation
practices, which leverage $\kappa$ on pairs of coders
\cite{artstein2008inter}.

The \F measures show that the four annotators reached a high level of
agreement for span annotations. As indicated by the Exact-F$_1$, their
intuitions were more consistent when labelling experiencers than
events (F$_1$=.86 and .34, respectively).  In fact, when considering
the Relaxed variant, agreement relative to experiencers does not
increase much (.88), while that concerning events is more than doubled
(.86). This shows that annotators agree on the event with a token
overlap, but do not exactly agree on the exact span.  Also Cohen's
$\kappa$ points to a good agreement for the span annotations.

We further observe a moderate emotion agreement ($\kappa$=.62),
calculated on all experiencer annotations with an overlap between
annotators. We break down emotion-class-specific agreement in
Table~\ref{tab:iaa-emo}. To give a better idea of their relative
occurrences, we indicate the counts of writer (column \#~Writer) or
another mentioned entity (\#~Other) being associated to each emotion.
We see that the agreement tends to be low only for particularly
infrequent emotion classes (e.g., \textit{pride},
\textit{contentment}), with \textit{surprise} being an exception to
this general pattern.

\begin{table}
  \centering\small
  \begin{tabular}{lrrr}
    \toprule
    Emotion Class  &  Cohen's $\kappa$ & \# Writer & \# Other \\
    \cmidrule(r){1-1}\cmidrule(rl){2-2}\cmidrule(rl){3-3}\cmidrule(l){4-4}
    Anger          & 0.53 & 204& 132\\
    Anticipation   & 0.58& 0 & 2\\
    Contentment    & 0.00& 2& 3\\
    Disappointment & 0.58 & 2& 4\\
    Disgust 	   & 0.80& 66& 21\\
    Fear           & 0.71& 134& 86 \\
    Frustration    & 0.09& 3& 2\\
    Guilt	   & 0.68 & 164& 95\\
    Hope           & 0.17 & 9& 30\\
    Joy            & 0.84 & 116& 146 \\
    Pride          & 0.00& 0& 1\\
    Sadness        & 0.62& 243& 170\\
    Shame          & 0.43&81 & 24\\
    Surprise       & 0.02&48& 21\\
    Trust          & 0.00& 0& 4\\
    No emotion     & 0.43 & 42& 227\\
    Other          & 0.17 & 4& 3\\
    \bottomrule
  \end{tabular}
  \caption{Inter-annotator agreement for separate emotions, calculated
    on all experiencer spans in which at least two annotators agreed
    by at least one token. \#~Writer and \#~Other denote the number of
    times in which either experiencer has such emotions.}
  \label{tab:iaa-emo}
\end{table}

As for appraisals, we analyze agreement based on the two measures in
Table~\ref{tab:iaa-appraisals}. The Cohen's $\kappa$ score is based on
a binarization based on the threshold of $\geq 4$ for an appraisal
category to hold. We do that under the assumption that some users of
our corpus might prefer to perform discrete categorization instead of
regression. In addition, we report the Spearman's correlation score on
the original, non-binarized ratings.  Note that not all annotators
might have marked the same event participants. Similar to emotions, we
compute their agreement only for sentences in which there is an
experiencer overlap. We see that the agreement is moderate to good
across nearly all dimensions. The judgments are positively correlated
across all dimensions, with the weakest correlations holding for
\effort, \attention and \urgent.

\begin{table}
  \centering\small
  \begin{tabular}{lrr}
    \toprule
    Variable: Appraisals & Cohen's $\kappa$ & Spearman's $\rho$\\
    \cmidrule(r){1-1}\cmidrule(rl){2-2}\cmidrule(l){3-3}
    Suddenness             &  0.62 & 0.51 \\
    Familiarity            &  0.53 & 0.35 \\
    Pleasantness           &  0.82 & 0.69 \\
    Understand             &  0.88 & 0.34 \\
    Goal relevance         &  0.57 & 0.43 \\
    Self responsibility    &  0.80 & 0.79 \\
    other responsibility   &  0.77 & 0.77 \\
    Situational respons.   &  0.66 & 0.61 \\
    Effort                 &  0.56 & 0.53 \\
    Exert                  &  0.52 & 0.20 \\
    Attend                 &  0.53 & 0.25 \\
    Consider               &  0.56 & 0.49 \\
    Outcome probability    &  0.63 & 0.48 \\
    Expectation discrepancy&  0.68 & 0.55 \\
    Goal conduciveness     &  0.68 & 0.62 \\
    Urgency                &  0.45 & 0.24 \\
    Self control           &  0.61 & 0.52 \\
    Other control          &  0.67 & 0.60 \\
    Situational control    &  0.60 & 0.53 \\
    Adjustment check       &  0.69 & 0.56 \\
    Internal check         &  0.56 & 0.52 \\
    External check         &  0.67 & 0.65 \\
    \bottomrule
  \end{tabular}
  \caption{Agreement on appraisals. $\kappa$
    is based on a discretization at the threshold $\geq 4$ to two
    classes.}
  \label{tab:iaa-appraisals}
\end{table}

\subsection{Aggregated Corpus Analyses}
The final corpus encompasses 720 texts (929 sentences).\footnote{We
  release 180 extra texts that were not annotated by all
  annotators. For simplicity, we do not discuss them here.}  It
includes 912 event spans, on a total of 17.5k tokens (on average, 24.3
per sentence), and contains annotations for 1329
experiencers.\footnote{Word and sentence counts were obtained using
  the \textsc{nltk} tokenizer. Prior to tokenization, we removed the
  ``\ldots'' masks.} Experiencers are mostly a combination of writer
and other entities within a sentence (in 270 texts, the experiencer is
the writer only; in 8, only other entities). Further, the predominant
emotions associated to the writers (\anger, \guilt, \fear, \joy,
\shame, \disgust, \sadness, see Table~\ref{tab:iaa-emo}) correspond to
the set of emotions for which the authors of texts in en\textsc{Isear}
self-labelled their own reactions.

\subsubsection{Within Experiencer Analysis}
\label{sub:within-exp}
\begin{figure*}
  \centering
  \includegraphics[width=.9\linewidth]{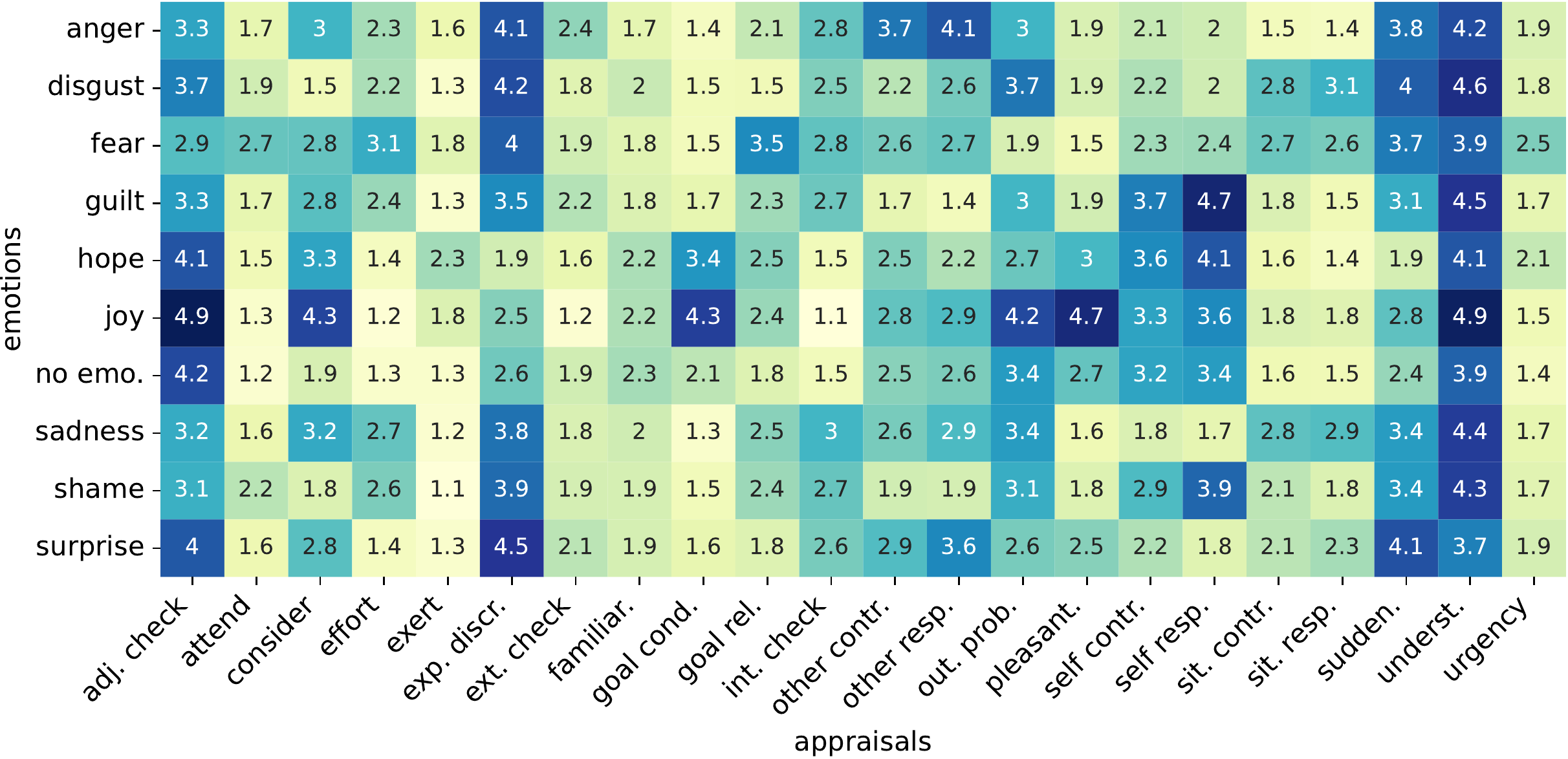}
  \caption{Analysis of appraisals and emotions for each experiencer.}
  \label{fig:emo_appr_avgs}
\end{figure*}
We now turn to the analysis of the emotion and appraisal annotations
for each experiencer in isolation, illustrated in
Figure~\ref{fig:emo_appr_avgs}. The heatmap shows the average
appraisal values for each emotion (limited to those that appear in the
corpus more often than 15 times).

Some appraisal dimensions have similar average scores across emotion
categories. This is the case for appraisals that resulted in
acceptable $\kappa$ agreement scores, but seemed particularly
difficult to annotate during guidelines discussion. Columns that stand
out are \adjCheck, \discrepancy and \understand, which tend to be
higher than the others.  For instance, while \understand has lower
averages for \textit{fear} and \textit{surprise}, it does not show
high variability across cells, suggesting that evaluating in
retrospect whether an experiencer understood what was going on during
the event is difficult. A similar consideration holds for
\textit{effort} and \textit{exert}.

By contrast, many appraisal dimensions hold more for specific
emotions, like \selfControl and \selfResp, which mainly concentrate on
the rows from \textit{guilt} to \textit{no emotion}.  Some intuitive
features of emotions emerge there, indicating that certain categories
are more strongly characterized by specific appraisal dimensions than
others.  For instance, the label \textit{surprise} seems to be chosen
when events are appraised as sudden (\suddenness) or as divergent from
one's expectation (\discrepancy), and anger is more typical to events
triggered by others (\otherResp) instead of the person reporting on
it.  The dimension of \selfControl is particularly high for
\textit{guilt}, among the set of negatively-valenced emotions, but not
for its kin \textit{shame}. \textit{Joy} and \hope, the only positive
emotions in the table, exhibit a greater score of \pleasantness. They
are also associated to a particularly high level of \adjCheck (i.e.,
the idea that one can easily cope with the consequences of the event
and \textit{outcome probability} (the ability to predict the
consequence of the event) as opposed to \fear.

\subsubsection{Between-Experiencer Analysis}
The novelty of \corpusname is that we can evaluate the
relation of emotions and appraisals between different experiencers. We
look into that in the following.

\begin{figure}
  \centering
  \includegraphics[width=0.9\linewidth]{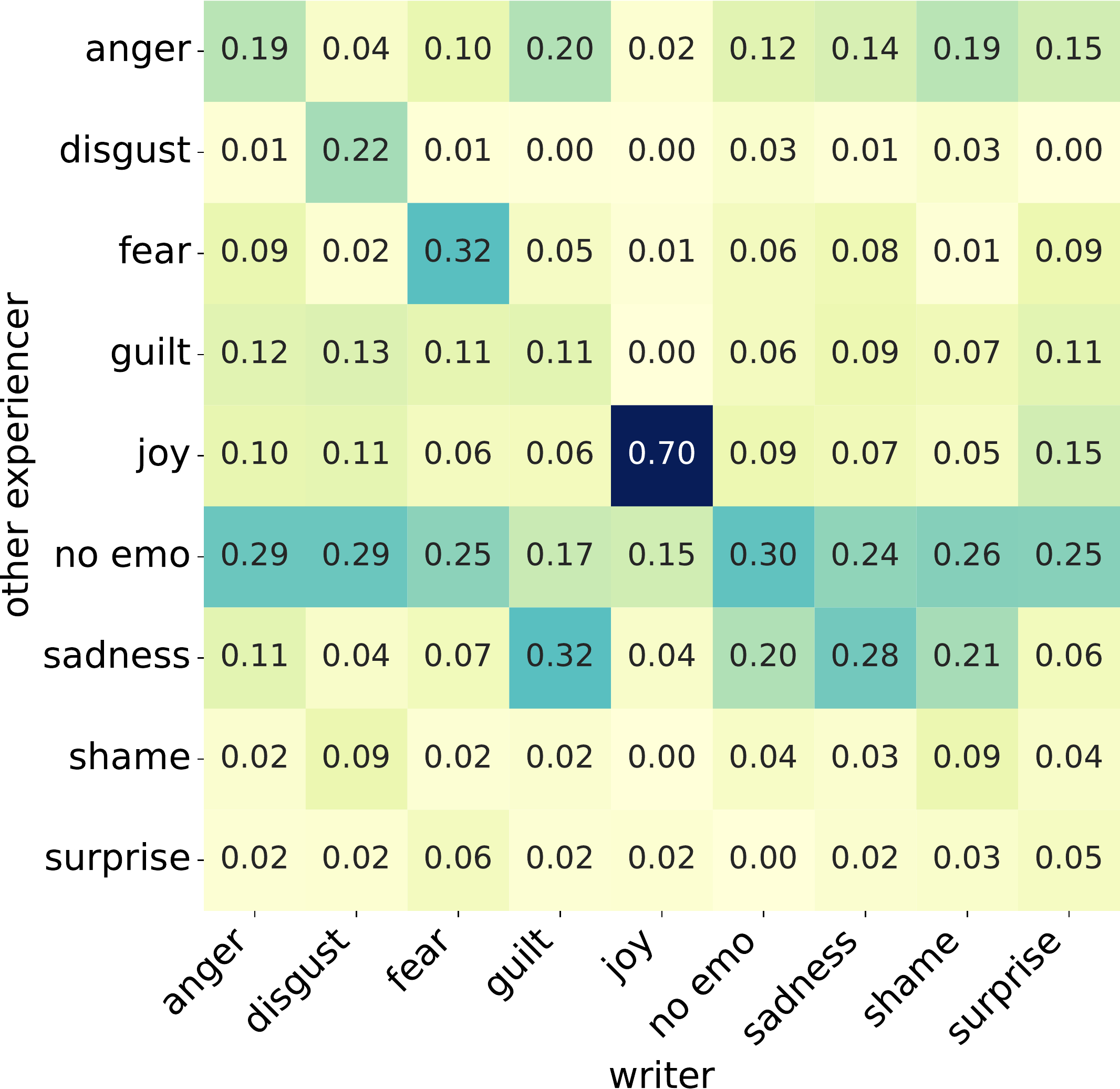}
  \caption{Averaged emotion co-occurrences between the writer
    (columns) and other experiencers (rows).}
  \label{fig:emo_co_occ}
\end{figure}

\paragraph{Different Experiencers, Different Reactions.}
If one experiencer feels a given emotion in response to an event, what
emotions can be elicited in the other participants? We analyse the
relation between the emotion of the writer vs.\ that of any other
experiencer, and do that for the texts in which both appear -- i.e.,
ignoring instances where only one of them is annotated. Results are
shown in Table~\ref{fig:emo_co_occ}, where one cell represents the
proportion of times any experiencer is associated to the emotion on
the column, when the writer is annotated with the emotion on the
row. We disregard infrequent classes in this depiction.

\begin{figure*}
  \centering
  \includegraphics[width=0.75\linewidth]{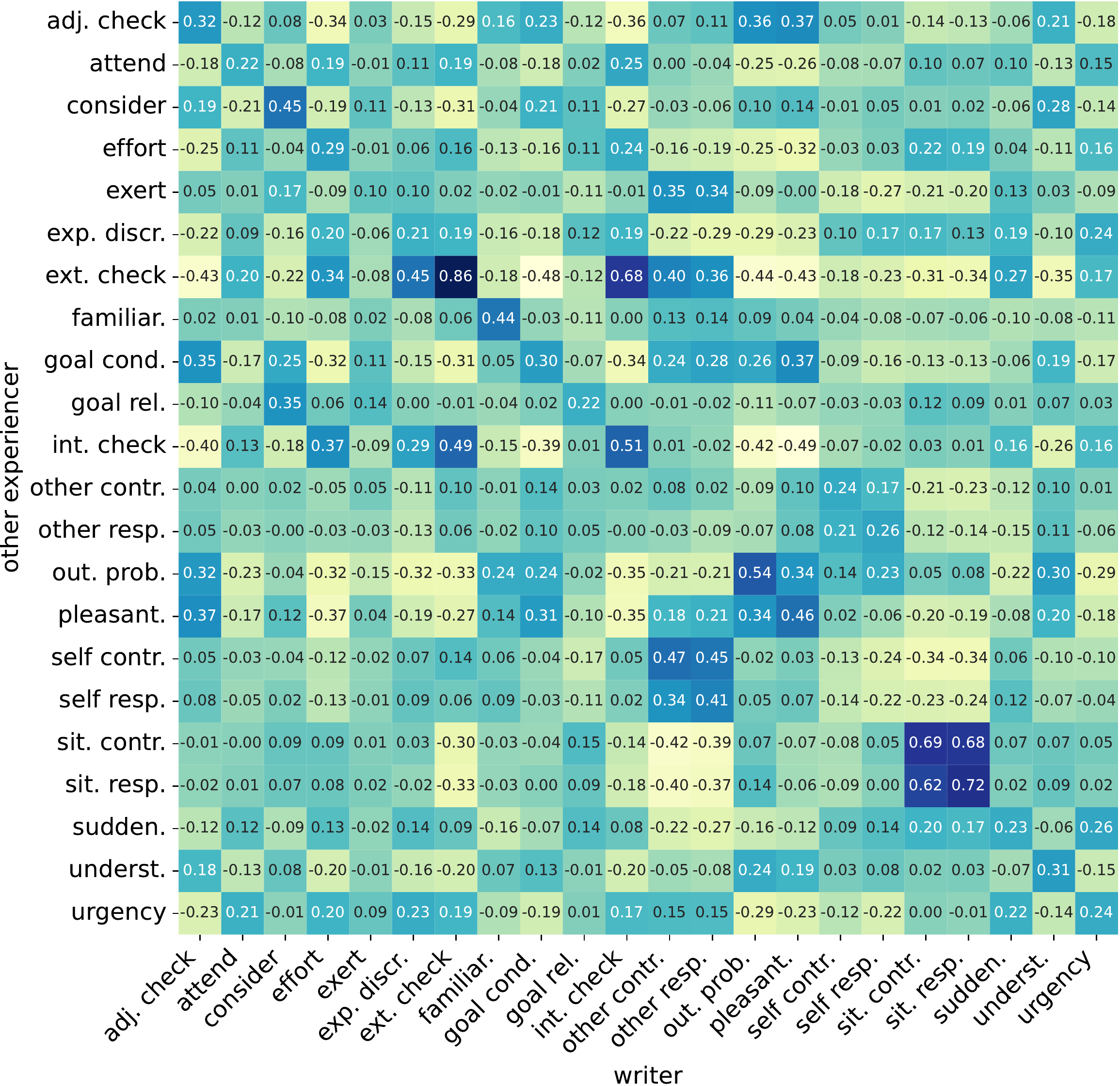}
  \caption{Spearman's correlations of the writer's (columns) and other
    experiencers' (rows) appraisal scores.}
  \label{fig:appr_corr}
\end{figure*}

The class joy is standing out, as it seems shared by all participants in an
event. The diagonal contains comparably high values for some other classes as
well. This means that in some cases an emotion is common to different
experiencers. However, for the majority of classes, the higher numbers are
scattered off diagonal. That is, different emotion reactions can be inferred
from text with respect to different semantic roles.  Interesting combinations
are \textit{guilt}--\textit{anger} (.20) \textit{no emotion}--\textit{sadness}
(.20).  Next to that, the combination of \textit{guilt} and \textit{sadness}
(.32) seems to suggest that the writer's sadness is often accompanied by
another's guilt. Another interesting case is in the writer's emotion of
\textit{shame}, which often co-occurs with the \textit{anger} of other entities
(.19). Lastly, it is worth noticing that other experiencers often cannot
attribute emotions to a situation that, instead, caused some in the writer.
That happens for all of writer's emotions, with the lower cases being
\textit{joy} and \textit{guilt} -- i.e., two labels that, as discussed, likely
co-occur with another emotion. The opposite is not true: non-emotional
reactions of the writers are not uniformly distributed across the emotions of
other entities.

\paragraph{Appraisal Correlations.}
Next to emotions, we are interested in the relation between appraisals
across different entities: if one perceives, for instance, \selfResp
for an event, what is the appraisal of the other experiencer? Results
are in Figure~\ref{fig:appr_corr}.  For each (writer-other entity)
pair in a text, we retrieve the scores of all appraisal pairs, where
one element is the (averaged, see Section~\ref{sec:data-aggregation})
score assigned to the writer and the other is given to the mentioned
entity. Hence, we calculate Spearman's correlation for such appraisal
combinations corresponding to the cells of \autoref{fig:appr_corr}.

Appraisals holding for the writers are positively correlated with the
same dimensions for other entities (see diagonal). This, however, does
not suggest that events are always similarly appraised by all
participants, as many positive correlations can be found among diverse
appraisal combinations. Examples are \textit{internal
  check}-\textit{external check} ($\rho$=.68), \textit{expectation
  discrepancy}-\textit{external check} (.45), and \textit{other
  control} (or \textit{responsibility}) - \textit{self control}
(\textit{responsibility}).  The latter pair indicates that, often, one
participant triggers the event and the other is subject to it -- but
if an event is driven by external factors, it is so for both (see
their \textit{situational control/responsibility}).  Among the
negatively correlated, we notice \textit{internal
  check}-\textit{pleasantness}, \textit{outcome
  probability}-\textit{urgency},
\textit{external check}-\textit{adjustment check}.

\section{Conclusion}
This paper introduced \corpusname, an English dataset motivated by
cognitive appraisal theories of emotions and endowed with a
multi-level set of annotations, including all participants in an
emotion episode, their specific reaction, and their relation to the
eliciting events. Our analysis shows that the text-level emotion
annotation is typically not the same for all experiencers, hence,
modelling emotion classification and role labeling together provides a
more complete picture of emotion descriptions in text.

Our corpus is the fundament for future research to develop
experiencer-specific emotion analysis models. These will then
enable a large-scale analysis of causal chains of emotions in context
of multiple people.

\section*{Acknowledgements}
This research is funded by the German Research Council (DFG), project
``Computational Event Analysis based on Appraisal Theories for Emotion
Analysis'' (CEAT, project number KL 2869/1-2).  We thank our
annotators and Kai Sassenberg for fruitful discussions.

\clearpage

\section{Bibliographical References}\label{reference}
\label{main:ref}


\begin{thebibliography}{}

\bibitem[\protect\citename{Alm \bgroup et al.\egroup }2005]{Ovesdotter2005}
Alm, C.~O., Roth, D., and Sproat, R.
\newblock (2005).
\newblock Emotions from text: Machine learning for text-based emotion
  prediction.
\newblock In {\em Proceedings of Human Language Technology Conference and
  Conference on Empirical Methods in Natural Language Processing}, pages
  579--586, Vancouver, British Columbia, Canada. Association for Computational
  Linguistics.

\bibitem[\protect\citename{Aman and Szpakowicz}2007]{aman2007identifying}
Aman, S. and Szpakowicz, S.
\newblock (2007).
\newblock Identifying expressions of emotion in text.
\newblock In {\em International Conference on Text, Speech and Dialogue}, pages
  196--205. Springer.

\bibitem[\protect\citename{Artstein and Poesio}2008]{artstein2008inter}
Artstein, R. and Poesio, M.
\newblock (2008).
\newblock Inter-coder agreement for computational linguistics.
\newblock {\em Computational linguistics}, 34(4):555--596.

\bibitem[\protect\citename{Balahur \bgroup et al.\egroup
  }2011]{balahur2011building}
Balahur, A., Hermida, J.~M., and Montoyo, A.
\newblock (2011).
\newblock Building and exploiting emotinet, a knowledge base for emotion
  detection based on the appraisal theory model.
\newblock {\em IEEE transactions on affective computing}, 3(1):88--101.

\bibitem[\protect\citename{Bostan \bgroup et al.\egroup }2020]{Bostan2020}
Bostan, L. A.~M., Kim, E., and Klinger, R.
\newblock (2020).
\newblock {GoodNewsEveryone}: A corpus of news headlines annotated with
  emotions, semantic roles, and reader perception.
\newblock In Nicoletta Calzolari, et~al., editors, {\em Proceedings of the 12th
  International Conference on Language Resources and Evaluation (LREC'20)},
  Marseille, France. European Language Resources Association (ELRA).

\bibitem[\protect\citename{Buechel and Hahn}2017a]{Buechel2017}
Buechel, S. and Hahn, U.
\newblock (2017a).
\newblock {E}mo{B}ank: Studying the impact of annotation perspective and
  representation format on dimensional emotion analysis.
\newblock In {\em Proceedings of the 15th Conference of the {E}uropean Chapter
  of the Association for Computational Linguistics: Volume 2, Short Papers},
  pages 578--585, Valencia, Spain. Association for Computational Linguistics.

\bibitem[\protect\citename{Buechel and Hahn}2017b]{buechel-hahn-2017-readers}
Buechel, S. and Hahn, U.
\newblock (2017b).
\newblock Readers vs. writers vs. texts: Coping with different perspectives of
  text understanding in emotion annotation.
\newblock In {\em Proceedings of the 11th Linguistic Annotation Workshop},
  pages 1--12, Valencia, Spain, April. Association for Computational
  Linguistics.

\bibitem[\protect\citename{Cambria \bgroup et al.\egroup }2020]{Cambria2020}
Cambria, E., Li, Y., Xing, F.~Z., Poria, S., and Kwok, K.
\newblock (2020).
\newblock Senticnet 6: Ensemble application of symbolic and subsymbolic ai for
  sentiment analysis.
\newblock In {\em Proceedings of the 29th ACM International Conference on
  Information \& Knowledge Management}, CIKM '20, page 105–114, New York, NY,
  USA. Association for Computing Machinery.

\bibitem[\protect\citename{Casel \bgroup et al.\egroup }2021]{Casel2021}
Casel, F., Heindl, A., and Klinger, R.
\newblock (2021).
\newblock Emotion recognition under consideration of the emotion component
  process model.
\newblock In {\em Proceedings of the 17th Conference on Natural Language
  Processing (KONVENS 2021)}, pages 49--61, D{\"u}sseldorf, Germany, 6--9
  September. KONVENS 2021 Organizers.

\bibitem[\protect\citename{Chang \bgroup et al.\egroup
  }2015]{chang-etal-2015-linguistic}
Chang, Y.-C., Chen, C.-C., Hsieh, Y.-L., Chen, C.~C., and Hsu, W.-L.
\newblock (2015).
\newblock Linguistic template extraction for recognizing reader-emotion and
  emotional resonance writing assistance.
\newblock In {\em Proceedings of the 53rd Annual Meeting of the Association for
  Computational Linguistics and the 7th International Joint Conference on
  Natural Language Processing (Volume 2: Short Papers)}, pages 775--780,
  Beijing, China, July. Association for Computational Linguistics.

\bibitem[\protect\citename{Chen \bgroup et al.\egroup
  }2020]{chen-etal-2020-end}
Chen, Y., Hou, W., Li, S., Wu, C., and Zhang, X.
\newblock (2020).
\newblock End-to-end emotion-cause pair extraction with graph convolutional
  network.
\newblock In {\em Proceedings of the 28th International Conference on
  Computational Linguistics}, pages 198--207, Barcelona, Spain (Online),
  December. International Committee on Computational Linguistics.

\bibitem[\protect\citename{Cohen}1960]{cohen1960coefficient}
Cohen, J.
\newblock (1960).
\newblock A coefficient of agreement for nominal scales.
\newblock {\em Educational and psychological measurement}, 20(1):37--46.

\bibitem[\protect\citename{Demszky \bgroup et al.\egroup
  }2020]{demszky-etal-2020-goemotions}
Demszky, D., Movshovitz-Attias, D., Ko, J., Cowen, A., Nemade, G., and Ravi, S.
\newblock (2020).
\newblock {G}o{E}motions: A dataset of fine-grained emotions.
\newblock In {\em Proceedings of the 58th Annual Meeting of the Association for
  Computational Linguistics}, pages 4040--4054, Online, July. Association for
  Computational Linguistics.

\bibitem[\protect\citename{Ekman}1992]{Ekman1992}
Ekman, P.
\newblock (1992).
\newblock An argument for basic emotions.
\newblock {\em Cognition \& emotion}, 6(3-4):169--200.

\bibitem[\protect\citename{Ellsworth and Smith}1988]{Ellsworth1988}
Ellsworth, P.~C. and Smith, C.~A.
\newblock (1988).
\newblock From appraisal to emotion: Differences among unpleasant feelings.
\newblock {\em Motivation and emotion}, 12(3):271--302.

\bibitem[\protect\citename{Fillmore and others}1976]{fillmore1976frame}
Fillmore, C.~J. et~al.
\newblock (1976).
\newblock Frame semantics and the nature of language.
\newblock {\em Annals of the New York Academy of Sciences: Conference on the
  origin and development of language and speech}, 280(1):20--32.

\bibitem[\protect\citename{Gao \bgroup et al.\egroup }2017]{gao2017overview}
Gao, Q., Jiannan, H., Ruifeng, X., Lin, G., He, Y., Wong, K.-F., and Lu, Q.
\newblock (2017).
\newblock Overview of ntcir-13 eca task.
\newblock In {\em Proceedings of the NTCIR-13 Conference}.

\bibitem[\protect\citename{Ghazi \bgroup et al.\egroup
  }2015]{ghazi2015detecting}
Ghazi, D., Inkpen, D., and Szpakowicz, S.
\newblock (2015).
\newblock Detecting emotion stimuli in emotion-bearing sentences.
\newblock In {\em International Conference on Intelligent Text Processing and
  Computational Linguistics}, pages 152--165. Springer.

\bibitem[\protect\citename{Gui \bgroup et al.\egroup
  }2016]{gui-etal-2016-event}
Gui, L., Wu, D., Xu, R., Lu, Q., and Zhou, Y.
\newblock (2016).
\newblock Event-driven emotion cause extraction with corpus construction.
\newblock In {\em Proceedings of the 2016 Conference on Empirical Methods in
  Natural Language Processing}, pages 1639--1649, Austin, Texas, November.
  Association for Computational Linguistics.

\bibitem[\protect\citename{Haider \bgroup et al.\egroup }2020]{Haider2020}
Haider, T., Eger, S., Kim, E., Klinger, R., and Menninghaus, W.
\newblock (2020).
\newblock {PO-EMO}: Conceptualization, annotation, and modeling of aesthetic
  emotions in {German} and {English} poetry.
\newblock In Nicoletta Calzolari, et~al., editors, {\em Proceedings of the 12th
  International Conference on Language Resources and Evaluation (LREC'20)},
  Marseille, France, May. European Language Resources Association (ELRA).

\bibitem[\protect\citename{Hofmann \bgroup et al.\egroup }2020]{Hofmann2020}
Hofmann, J., Troiano, E., Sassenberg, K., and Klinger, R.
\newblock (2020).
\newblock Appraisal theories for emotion classification in text.
\newblock In {\em Proceedings of the 28th International Conference on
  Computational Linguistics}, pages 125--138, Barcelona, Spain (Online).
  International Committee on Computational Linguistics.

\bibitem[\protect\citename{Hofmann \bgroup et al.\egroup }2021]{Hofmann2021}
Hofmann, J., Troiano, E., and Klinger, R.
\newblock (2021).
\newblock Emotion-aware, emotion-agnostic, or automatic: Corpus creation
  strategies to obtain cognitive event appraisal annotations.
\newblock In {\em Proceedings of the Eleventh Workshop on Computational
  Approaches to Subjectivity, Sentiment and Social Media Analysis}, pages
  160--170, Online. Association for Computational Linguistics.

\bibitem[\protect\citename{Kessler \bgroup et al.\egroup }2010]{Kessler2010}
Kessler, J.~S., Eckert, M., Clark, L., and Nicolov, N.
\newblock (2010).
\newblock {The 2010 ICWSM JDPA Sentment Corpus for the Automotive Domain}.
\newblock In {\em 4th International AAAI Conference on Weblogs and Social Media
  Data Workshop Challenge (ICWSM-DWC 2010)}.

\bibitem[\protect\citename{Kim and Klinger}2018]{Kim2018}
Kim, E. and Klinger, R.
\newblock (2018).
\newblock Who feels what and why? annotation of a literature corpus with
  semantic roles of emotions.
\newblock In {\em Proceedings of the 27th International Conference on
  Computational Linguistics}, pages 1345--1359, Santa Fe, New Mexico, USA,
  August. Association for Computational Linguistics.

\bibitem[\protect\citename{Kim and Klinger}2019]{Kim2019}
Kim, E. and Klinger, R.
\newblock (2019).
\newblock Frowning {F}rodo, wincing {L}eia, and a seriously great friendship:
  Learning to classify emotional relationships of fictional characters.
\newblock In {\em Proceedings of the 2019 Conference of the North {A}merican
  Chapter of the Association for Computational Linguistics: Human Language
  Technologies, Volume 1 (Long and Short Papers)}, pages 647--653, Minneapolis,
  Minnesota, June. Association for Computational Linguistics.

\bibitem[\protect\citename{Klie \bgroup et al.\egroup }2018]{tubiblio106270}
Klie, J.-C., Bugert, M., Boullosa, B., de~Castilho, R.~E., and Gurevych, I.
\newblock (2018).
\newblock The inception platform: Machine-assisted and knowledge-oriented
  interactive annotation.
\newblock In {\em Proceedings of the 27th International Conference on
  Computational Linguistics: System Demonstrations}, pages 5--9. Association
  for Computational Linguistics, Juni.

\bibitem[\protect\citename{Klinger and Cimiano}2014]{Klinger2014}
Klinger, R. and Cimiano, P.
\newblock (2014).
\newblock The {USAGE} review corpus for fine grained multi lingual opinion
  analysis.
\newblock In {\em Proceedings of the Ninth International Conference on Language
  Resources and Evaluation ({LREC}'14)}, pages 2211--2218, Reykjavik, Iceland,
  May. European Language Resources Association (ELRA).

\bibitem[\protect\citename{Li and Xu}2014]{li2014text}
Li, W. and Xu, H.
\newblock (2014).
\newblock Text-based emotion classification using emotion cause extraction.
\newblock {\em Expert Systems with Applications}, 41(4):1742--1749.

\bibitem[\protect\citename{Mohammad \bgroup et al.\egroup }2014]{Mohammad2014}
Mohammad, S., Zhu, X., and Martin, J.
\newblock (2014).
\newblock Semantic role labeling of emotions in tweets.
\newblock In {\em Proceedings of the 5th Workshop on Computational Approaches
  to Subjectivity, Sentiment and Social Media Analysis}, pages 32--41,
  Baltimore, Maryland, June. Association for Computational Linguistics.

\bibitem[\protect\citename{Mohammad \bgroup et al.\egroup }2018]{Mohammad2018}
Mohammad, S., Bravo-Marquez, F., Salameh, M., and Kiritchenko, S.
\newblock (2018).
\newblock {S}em{E}val-2018 task 1: Affect in tweets.
\newblock In {\em Proceedings of The 12th International Workshop on Semantic
  Evaluation}, pages 1--17, New Orleans, Louisiana. Association for
  Computational Linguistics.

\bibitem[\protect\citename{Mohammad}2012]{mohammad-2012-emotional}
Mohammad, S.
\newblock (2012).
\newblock {\#}emotional tweets.
\newblock In {\em *{SEM} 2012: The First Joint Conference on Lexical and
  Computational Semantics {--} Volume 1: Proceedings of the main conference and
  the shared task, and Volume 2: Proceedings of the Sixth International
  Workshop on Semantic Evaluation ({S}em{E}val 2012)}, pages 246--255,
  Montr{\'e}al, Canada, 7-8 June. Association for Computational Linguistics.

\bibitem[\protect\citename{Neviarouskaya and Aono}2013]{Neviarouskaya2013}
Neviarouskaya, A. and Aono, M.
\newblock (2013).
\newblock Extracting causes of emotions from text.
\newblock In {\em Proceedings of the Sixth International Joint Conference on
  Natural Language Processing}, pages 932--936, Nagoya, Japan, October. Asian
  Federation of Natural Language Processing.

\bibitem[\protect\citename{Oberl{\"a}nder and Klinger}2020]{Oberlander2020}
Oberl{\"a}nder, L. A.~M. and Klinger, R.
\newblock (2020).
\newblock Token sequence labeling vs. clause classification for {E}nglish
  emotion stimulus detection.
\newblock In {\em Proceedings of the Ninth Joint Conference on Lexical and
  Computational Semantics}, pages 58--70, Barcelona, Spain (Online), December.
  Association for Computational Linguistics.

\bibitem[\protect\citename{Oberl\"ander \bgroup et al.\egroup
  }2020]{Oberlaender2020b}
Oberl\"ander, L., Reich, K., and Klinger, R.
\newblock (2020).
\newblock Experiencers, stimuli, or targets: Which semantic roles enable
  machine learning to infer the emotions?
\newblock In {\em Proceedings of the Third Workshop on Computational Modeling
  of People{'}s Opinions, Personality, and Emotions in Social Media},
  Barcelona, Spain, December. Association for Computational Linguistics.

\bibitem[\protect\citename{Omdahl}1995]{omdahl1995cognitive}
Omdahl, B.~L.
\newblock (1995).
\newblock {\em Cognitive Appraisal, Emotion, and Empathy}.
\newblock Mahwah, NJ: Lawrence Erlbaum.

\bibitem[\protect\citename{Plutchik}2001]{Plutchik2001}
Plutchik, R.
\newblock (2001).
\newblock The nature of emotions.
\newblock {\em American Scientist}, 89(4):344--350.

\bibitem[\protect\citename{Preo{\c{t}}iuc-Pietro \bgroup et al.\egroup
  }2016]{Preotiuc2016}
Preo{\c{t}}iuc-Pietro, D., Schwartz, H.~A., Park, G., Eichstaedt, J., Kern, M.,
  Ungar, L., and Shulman, E.
\newblock (2016).
\newblock Modelling valence and arousal in {F}acebook posts.
\newblock In {\em Proceedings of the 7th Workshop on Computational Approaches
  to Subjectivity, Sentiment and Social Media Analysis}, pages 9--15, San
  Diego, California. Association for Computational Linguistics.

\bibitem[\protect\citename{Rashkin \bgroup et al.\egroup
  }2018]{rashkin-etal-2018-event2mind}
Rashkin, H., Sap, M., Allaway, E., Smith, N.~A., and Choi, Y.
\newblock (2018).
\newblock {E}vent2{M}ind: Commonsense inference on events, intents, and
  reactions.
\newblock In {\em Proceedings of the 56th Annual Meeting of the Association for
  Computational Linguistics (Volume 1: Long Papers)}, pages 463--473,
  Melbourne, Australia, July. Association for Computational Linguistics.

\bibitem[\protect\citename{Rashkin \bgroup et al.\egroup
  }2019]{rashkin-etal-2019-towards}
Rashkin, H., Smith, E.~M., Li, M., and Boureau, Y.-L.
\newblock (2019).
\newblock Towards empathetic open-domain conversation models: A new benchmark
  and dataset.
\newblock In {\em Proceedings of the 57th Annual Meeting of the Association for
  Computational Linguistics}, pages 5370--5381, Florence, Italy, July.
  Association for Computational Linguistics.

\bibitem[\protect\citename{Roseman \bgroup et al.\egroup }1990]{Roseman1990}
Roseman, I.~J., Spindel, M.~S., and Jose, P.~E.
\newblock (1990).
\newblock Appraisals of emotion-eliciting events: Testing a theory of discrete
  emotions.
\newblock {\em Journal of Personality and Social Psychology}, 59(5):899--915.

\bibitem[\protect\citename{Russell and Mehrabian}1977]{russell1977evidence}
Russell, J.~A. and Mehrabian, A.
\newblock (1977).
\newblock Evidence for a three-factor theory of emotions.
\newblock {\em Journal of research in Personality}, 11(3):273--294.

\bibitem[\protect\citename{Russo \bgroup et al.\egroup
  }2011]{russo-etal-2011-emocause}
Russo, I., Caselli, T., Rubino, F., Boldrini, E., and Mart{\'\i}nez-Barco, P.
\newblock (2011).
\newblock {EMOC}ause: An easy-adaptable approach to extract emotion cause
  contexts.
\newblock In {\em Proceedings of the 2nd Workshop on Computational Approaches
  to Subjectivity and Sentiment Analysis ({WASSA} 2.011)}, pages 153--160,
  Portland, Oregon, June. Association for Computational Linguistics.

\bibitem[\protect\citename{Scherer and Fontaine}2013]{Scherer2013}
Scherer, K.~R. and Fontaine, J.~J.
\newblock (2013).
\newblock Driving the emotion process: The appraisal component.
\newblock In J.~J.~R. Fontaine, et~al., editors, {\em Series in affective
  science. Components of emotional meaning: A sourcebook}, chapter~12, pages
  266--290. Oxford University Press, Oxford.

\bibitem[\protect\citename{Scherer and Wallbott}1994]{scherer1994evidence}
Scherer, K.~R. and Wallbott, H.~G.
\newblock (1994).
\newblock Evidence for universality and cultural variation of differential
  emotion response patterning.
\newblock {\em Journal of personality and social psychology}, 66(2):310.

\bibitem[\protect\citename{Scherer and Wallbott}1997]{Scherer1997}
Scherer, K.~R. and Wallbott, H.~G.
\newblock (1997).
\newblock The {ISEAR} questionnaire and codebook.
\newblock Geneva Emotion Research Group.

\bibitem[\protect\citename{Scherer}1989]{scherer1989appraisal}
Scherer, K.~R.
\newblock (1989).
\newblock Appraisal theory.
\newblock {\em Handbook of Cognition and Emotion}.

\bibitem[\protect\citename{Scherer}2005]{Scherer2005}
Scherer, K.~R.
\newblock (2005).
\newblock What are emotions? {And} how can they be measured?
\newblock {\em Social Science Information}, 44(4):695--729.

\bibitem[\protect\citename{Shaikh \bgroup et al.\egroup }2009]{Shaikh2009}
Shaikh, M. A.~M., Prendinger, H., and Ishizuka, M.
\newblock (2009).
\newblock A linguistic interpretation of the occ emotion model for affect
  sensing from text.
\newblock {\em Affective Information Processing}, pages 45--73.

\bibitem[\protect\citename{Smith and Ellsworth}1985]{Smith1985}
Smith, C.~A. and Ellsworth, P.~C.
\newblock (1985).
\newblock Patterns of cognitive appraisal in emotion.
\newblock {\em Journal of personality and social psychology}, 48(4):186--209.

\bibitem[\protect\citename{Toprak \bgroup et al.\egroup }2010]{Toprak2010}
Toprak, C., Jakob, N., and Gurevych, I.
\newblock (2010).
\newblock Sentence and expression level annotation of opinions in
  user-generated discourse.
\newblock In {\em Proceedings of the 48th Annual Meeting of the Association for
  Computational Linguistics}, pages 575--584, Uppsala, Sweden, July.
  Association for Computational Linguistics.

\bibitem[\protect\citename{Tracy and Robins}2006]{Tracy2006}
Tracy, J.~L. and Robins, R.~W.
\newblock (2006).
\newblock Appraisal antecedents of shame and guilt: Support for a theoretical
  model.
\newblock {\em Personality and social psychology bulletin}, 32(10):1339--1351.

\bibitem[\protect\citename{Troiano \bgroup et al.\egroup }2019]{Troiano2019}
Troiano, E., Pad{\'o}, S., and Klinger, R.
\newblock (2019).
\newblock Crowdsourcing and validating event-focused emotion corpora for
  {G}erman and {E}nglish.
\newblock In {\em Proceedings of the 57th Annual Meeting of the Association for
  Computational Linguistics}, pages 4005--4011, Florence, Italy. Association
  for Computational Linguistics.

\bibitem[\protect\citename{Wei \bgroup et al.\egroup
  }2020]{wei-etal-2020-effective}
Wei, P., Zhao, J., and Mao, W.
\newblock (2020).
\newblock Effective inter-clause modeling for end-to-end emotion-cause pair
  extraction.
\newblock In {\em Proceedings of the 58th Annual Meeting of the Association for
  Computational Linguistics}, pages 3171--3181, Online, July. Association for
  Computational Linguistics.

\bibitem[\protect\citename{Xia and Ding}2019]{xia-ding-2019-emotion}
Xia, R. and Ding, Z.
\newblock (2019).
\newblock Emotion-cause pair extraction: A new task to emotion analysis in
  texts.
\newblock In {\em Proceedings of the 57th Annual Meeting of the Association for
  Computational Linguistics}, pages 1003--1012, Florence, Italy, July.
  Association for Computational Linguistics.

\bibitem[\protect\citename{Yanchus}2006]{yanchus2006development}
Yanchus, N.~J.
\newblock (2006).
\newblock {\em Development and validation of a self-report cognitive appraisal
  scale}.
\newblock {Ph.D.} thesis, University of Georgia.

\bibitem[\protect\citename{Yu \bgroup et al.\egroup }2016]{Yu2016}
Yu, L.-C., Lee, L.-H., Hao, S., Wang, J., He, Y., Hu, J., Lai, K.~R., and
  Zhang, X.
\newblock (2016).
\newblock Building {C}hinese affective resources in valence-arousal dimensions.
\newblock In {\em Proceedings of the 2016 Conference of the North {A}merican
  Chapter of the Association for Computational Linguistics: Human Language
  Technologies}, pages 540--545, San Diego, California, June. Association for
  Computational Linguistics.

\end{thebibliography}
\end{document}